\documentclass[conference]{IEEEtran}
\IEEEoverridecommandlockouts
\usepackage{cite}
\usepackage{amsmath,amssymb,amsfonts}
\usepackage{algorithmic}
\usepackage{algorithm}
\usepackage{array}
\usepackage[caption=false,font=normalsize,labelfont=sf,textfont=sf]{subfig}
\usepackage{stfloats}
\usepackage{url}
\usepackage{verbatim}
\usepackage{cite}
\usepackage{graphicx}
\usepackage{textcomp}
\usepackage{xcolor}
\def\BibTeX{{\rm B\kern-.05em{\sc i\kern-.025em b}\kern-.08em
    T\kern-.1667em\lower.7ex\hbox{E}\kern-.125emX}}
\begin{document}

\title{Sparse Federated Training of Object Detection in the Internet of Vehicles
\thanks{This work is partially supported by Youth Foundation Project of Zhejiang Lab (No. K2023PD0AA01), partially supported by Research Initiation Project of Zhejiang Lab (No. 2022PD0AC02), partially supported the National Natural Science Foundation of China under Grant No. 62002170 and 62071222 (Corresponding author: Chuan Ma).}
}

\author{\IEEEauthorblockN{Luping Rao\IEEEauthorrefmark{1}, Chuan Ma\IEEEauthorrefmark{2}, Ming Ding\IEEEauthorrefmark{3}, Yuwen Qian\IEEEauthorrefmark{1}, Lu Zhou\IEEEauthorrefmark{4}, Zhe Liu\IEEEauthorrefmark{2}}
\IEEEauthorblockA{\IEEEauthorrefmark{1}Nanjing University of Science and Technology, Nanjing, China\\\IEEEauthorrefmark{2}Zhejiang Lab, Hang Zhou, China \\\IEEEauthorrefmark{3}Data61, CSIRO, Sydney, Australia\\\IEEEauthorrefmark{4}Nanjing University of Aeronautics and Astronautics, Nanjing, China\\lupingrao@njust.edu.cn, chuan.ma@zhejianglab.edu.cn}}

\maketitle

\begin{abstract}
As an essential component part of the Intelligent Transportation System (ITS), the Internet of Vehicles (IoV) plays a vital role in alleviating traffic issues. Object detection is one of the key technologies in the IoV, which has been widely used to provide traffic management services by analyzing timely and sensitive vehicle-related information. However, the current object detection methods are mostly based on centralized deep training, that is, the sensitive data obtained by edge devices need to be uploaded to the server, which raises privacy concerns. To mitigate such privacy leakage, we first propose a federated learning-based framework, where well-trained local models are shared in the central server. However, since edge devices usually have limited computing power, plus a strict requirement of low latency in IoVs, we further propose a sparse training process on edge devices, which can effectively lighten the model, and ensure its training efficiency on edge devices, thereby reducing communication overheads. In addition, due to the diverse computing capabilities and dynamic environment, different sparsity rates are applied to edge devices. To further guarantee the performance, we propose, FedWeg, an improved aggregation scheme based on FedAvg, which is designed by the inverse ratio of sparsity rates. Experiments on the real-life dataset using YOLO show that the proposed scheme can achieve the required object detection rate while saving considerable communication costs. 
\end{abstract}


\section{Introduction}
In the process of building smart cities, the intelligent transportation system (ITS) plays a vital role in public transportation and safety management\cite{chen2020edge}, thus further accelerating the development of cutting-edge technologies and the industrial revolution. This makes it possible to use advanced communication technology to solve traffic problems. Among them, The Internet of Vehicles (IoV)\cite{cheng2020accessibility} is a typical one, which uses sensor technology to collect and process the status information of vehicles, then, according to different functional requirements, effectively guide and supervise the vehicles. In addition, with the help of artificial intelligence (AI) technologies, such as deep learning and computer vision, the modern IoV can also make wise decisions for drivers.

Object detection is one of the key technologies in the IoV, which is widely used in the field of intelligent monitoring and automatic driving. Object detection technology mainly realizes the requirements of some scenes through the collected vehicle speed, traffic flow, etc., where two common methods, image processing, and deep learning, are applied. With the rapid development of deep learning and improvement of equipment capability, the method based on deep learning raises consideration of both real-time and accuracy in several scenarios. However, the current mainstream object detection algorithms are based on centralized learning, in which all the collected data from the edge devices should be gathered in one center before training. In such a process, the timely and sensitive data will induce communication overheads, long delays, and potential privacy issues.

To address the mentioned issues, the research community has introduced federated learning\cite{konevcny2016federated} to allocate the process of training models on edge devices\cite{li2020federated}, then only model updates instead of raw data are uploaded. Compared with centralized training, although such decentralization can partially alleviate the transmission overhead, the large scale of neural networks for object detection cannot be directly deployed on edge devices, leading to a low convergence rate. Therefore, the over-large model should be optimized for edge devices. To improve the training efficiency and alleviate the transmission burden, various model compression-related algorithms, such as value quantization\cite{han2015deep}, model distillation\cite{hinton2015distilling}, sparsity\cite{liu2017learning} and low-rank decomposition\cite{peng2012rasl}, are proposed. For example, Wu \textit{et al.} \cite{DBLP:journals/corr/abs-2108-13323} proposed an adaptive framework of knowledge distillation between student and teacher models, in order to reduce the communication cost of the federated learning process. In \cite{DBLP:conf/sensys/0005SZZLC21}, the FedMask framework realized the efficient running of personalized and structured sparse CNN models on each device. In addition, taking into account the problem of client data heterogeneity, the authors in \cite{dai2022dispfl} proposed a personalized FL which customizes sparse local models for each client. 

However, there are few similar studies in the IoV. Therefore, we propose a sparse federated training scheme of object detection in the IoV. The main contributions of this work are listed as follows.
 \begin{enumerate}
\item{We propose a sparse federated learning-based framework to protect data privacy, and only lightweight models are transmitted and shared.}
\item{We improve the traditional average aggregation algorithm based on the inverse ratio of sparsity rates to further adapt to the dynamic sparsity rates on different edge devices.}
\item{We have conducted experiments on real-life datasets and verified that our scheme can achieve the required object detection rate while saving a considerable communication cost.}
\end{enumerate}

The rest structure of the paper is as follows. Section \uppercase\expandafter{\romannumeral2} briefly introduces the relevant algorithms used. Section \uppercase\expandafter{\romannumeral3} focuses on the proposed scheme and aggregation algorithm. Section \uppercase\expandafter{\romannumeral4} presents the experiments and analysis, and Section \uppercase\expandafter{\romannumeral5} gives a summary of the full text.

\section{Brief Description of Relevant Approaches}
This chapter briefly introduces federated learning, sparse training, and object detection algorithms.

\subsection{Federated Learning}
Federated learning is a distributed machine learning framework, which was first proposed in 2016\cite{konevcny2016federated}. Its core idea is that the model is trained in a decentralized way, and the model parameters are shared with the server without sharing any data. Therefore, the local data will not be uploaded to the server, which realizes data privacy protection. The purpose of sharing model parameters is to aggregate them on the server side and update the model. And the federated learning framework usually uses the federated averaging (FedAvg) algorithm\cite{mcmahan2017communication} to train the global model. The training process of FedAvg is as follows.

\subsubsection{Initialization }
Firstly, let $ \mathcal{K}$ represent the client set, and $\mathcal{D}=\left \{ D_{1} ,D_{2},\cdot\cdot\cdot,D_{k}\right\} $ represent the data set owned by each client. The total amount of data set is $\mathcal{N}$, which is divided into $\left\{ N_{1} ,N_{2},\cdot\cdot\cdot,N_{k}\right\}$. Then, in each communication round \(t\), the server will select a batch of clients $\mathcal{K}_{t}$ ($\mathcal{K}_{t}  \subseteq \mathcal{K}$) from the client set for training at random, then send the initialized global model parameters $w^{t}$ to each selected client \(k\) ($k\in \mathcal{K}_{t} $).

\subsubsection{Local Training}
Each client uses the local dataset $D_{k}$ to train the received model with \(E\) epochs. The purpose of each client is to minimize the following objective function:
\begin{equation}
 \min_{w_{k} \in R} L_{k}\left ( w_{k}  \right )=\frac{1 }{\left | D_{k} \right | } \sum_{\left ( x_{i},y_{i}\in D_{k}\right )}l_{i}\left ( y_{i}, f_{k}\left ( x_{i} ;w_{k}\right )\right ),
\end{equation}
where $D_{k}$ represents the local data containing input-output vector pairs $\left( x_{i},y_{i}\right )$, and $x_{i}, y_{i}\in R$, $w_{k}$ is the weight parameter of local model $f_{k}$, and $l_{i}\left ( \cdot  \right )$ denotes the local loss function. Then each client uploads model updates to the server.

\subsubsection{Average Aggregation} The server aggregates local model updates using the FedAvg algorithm to obtain a new round of iterative global model $w^{t+1}$. The aggregation process is as follows:
\begin{equation}
w^{t+1} =  \sum_{k\in \mathcal{K}_{t}} \frac{N_{k}}{\mathcal{N}}w_{k}^{t}.
\end{equation}

The server sends the updated parameters $w^{t+1}$ to the clients and repeats the above process until convergence.

\subsection{Sparse Training}
The purpose of sparse training is to generate sparse networks. Nowadays, model compression is widely used in edge devices with limited resources. It includes typical methods, such as model distillation\cite{hinton2015distilling}, value quantification\cite{han2015deep}, and sparsification\cite{liu2017learning}. Model distillation refers to knowledge distillation, which aims to transfer the knowledge learned from one or more large models to a lightweight model for easy deployment. Value quantization encodes the values (weights or activations, etc.) in the network with a low accuracy, such as converting a 32-bit floating point number to an 8-bit fixed-point number to reduce the size of the model. Sparsification is also called pruning, which removes a large number of redundant variables and retains only the most relevant explanatory variables. 

In addition, sparsification can be implemented at several levels, such as the model parameter level, channel level, or layer level\cite{liu2017learning}. Channel-level sparsity is flexible and easy to implement, and can usually be used in any traditional CNN network. In this paper, we design the sparsity at the channel level and further improve its efficiency.

\subsection{Object Detection Algorithm}
In IoV, YOLO (You Only Look Once) \cite{1}, as an excellent object detection algorithm, is widely used in vehicle, pedestrian detection, and other real-time tasks. The YOLO algorithm is a typical one-stage method, which means that the neural network only needs to look at the picture once to output the result. It was proposed by Joseph Redmon \textit{et al}. in 2016. So far, the YOLO algorithm has been developed to YOLOv7, of which YOLOv1 has laid the foundation of the entire YOLO series, and the subsequent YOLO algorithms have been continuously improved and innovated. In this paper, we use YOLOv3\cite{redmon2018yolov3} in the experiment, because compared with YOLOv1 and YOLOv2, the performance of YOLOv3 has been greatly improved, and the network structure is relatively complete, while the later version is only improved on the basis of YOLOv3.

YOLOv3 uses DarkNet-53 (Contains 53 convolutional layers) as the backbone network. The classification accuracy of Darknet-53 is comparable to ResNet-101 and ResNet-152, but the speed is much faster and the number of network layers is much less. The network draws on the residual network structure to form a deeper network level, as well as multi-scale detection, which improves the detection accuracy of small-size objects. 

\section{Object Detection Algorithm Based on \\ Yolo and Federated Learning}
In this section, we will present the details of the proposed sparse federated training framework and the FedWeg aggregation algorithm based on the FedAvg algorithm.

\subsection{Object Detection Based on Sparse Federated Training}

\begin{figure}
    \centering
    \includegraphics[width=\linewidth]{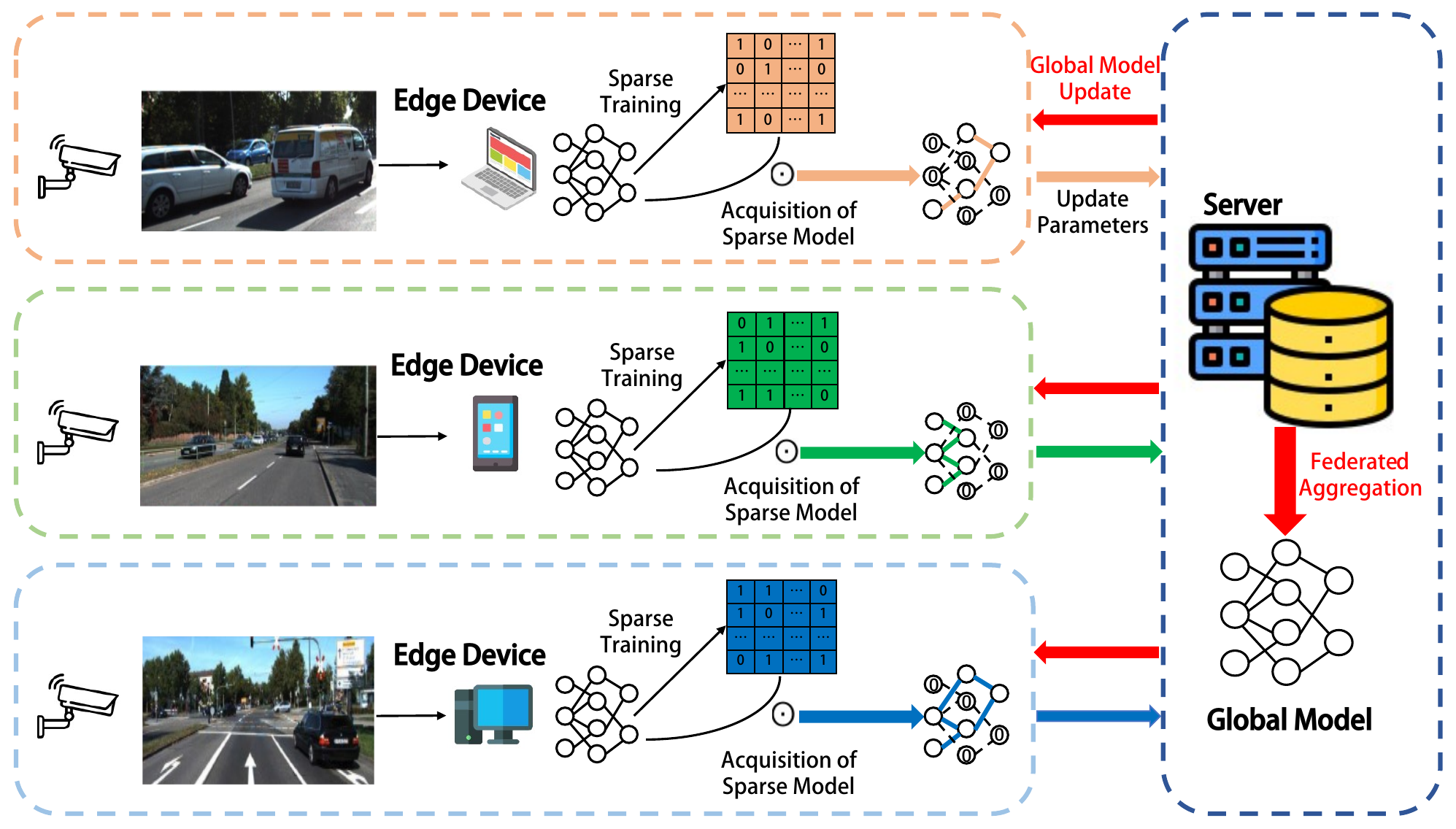}
    \caption{Sparse federated learning framework.}
    \label{fig:my_label}
\end{figure}

Considering the application requirements of actual scenarios, object detection is required, especially in the IoV scenarios, to achieve fast and accurate results. Inspired by federated learning and the sparse method, we design a sparse federated training framework, which can apply the sparse object detection model to edge devices. The specific process is divided into the following three steps. Firstly, the server initializes the object detection model and sends it to the edge devices. Secondly, the edge devices use a local dataset for sparse local training. Finally, after the model sparsification, the edge devices upload the model updates to the server for aggregation, and then the server distributes the updated model to the edge devices for a new communication round. An illustration is provided in Figure 1.

\subsubsection{Object Detection Model}
The server takes an initialized object detection model and distributes it to edge devices. YOLO is currently a fast object detection algorithm, which meets the requirements of real-time detection in the IoV. Then, from YOLO, the input image of the specified size is divided into \( S \) × \( S \) grids. Wherein each grid cell generates \( B \) boundary boxes and calculates the confidence score of the probability of the object in its corresponding \( B \) boundary boxes. Therefore, each bounding box is composed of five numbers: the coordinates \( x \) and \( y \) of the center point, the height \( h \) and width \( w \) of the normalized bounding box compared with the original image, and the confidence score \( C \). Then, the box with the largest IoU (the ratio between the intersection of two regions and the union of two regions) in the \( B \) boundary boxes is used to fit the ground truth, and the prediction box is finally obtained after a series of fine adjustments. The non-maximum suppression (NMS) method is used to accelerate the fitting process. We define the specific loss function. It includes coordinate loss ($L_{\mathrm{coo}}$), confidence score loss ($L_{\mathrm{con}}$), classification loss ($L_{\mathrm{cls}}$). The loss function is as follows:

\begin{equation}
L=\lambda_{\mathrm{coo}}L_{\mathrm{coo}}+\lambda_{\mathrm{cls}}L_{\mathrm{cls}}+ \lambda_{\mathrm{con}}L_{\mathrm{con}},
\end{equation}
where $ \lambda $ refers to each hyperparameter, and various loss functions are as follows:

\begin{equation}
L_{\mathrm{cls}}= BCE\left ( P,\hat{P}  \right );
\end{equation}

\begin{equation}
\begin{split}
L_{\mathrm{coo}}=\sum_{i=0}^{S^{2}}\phi_{i}^{\ast}\left [ \left(x_{i} -\hat{x}_{i}\right )^{2} + \left (y_{i} -\hat{y}_{i}\right )^{2} +\right. \\ \left.\left( \sqrt{w_{i}} -\sqrt{\hat{w}_{i}} \right ) ^{2}+\left (\sqrt{h_{i}}-\sqrt{\hat{h}_{i}} \right )^{2}\right];
\end{split}
\end{equation}

\begin{equation}
L_{\mathrm{con}}=\sum_{i=0}^{S^{2}}\phi_{i}^{\ast } \left (C_{O_{i} }- \hat{C} _{O_{i} }\right )^{2}, 
\end{equation}
where $\phi_{i}^{\ast}$ is the indicator function if the object appears in the cell \(i\), then $\phi_{i}^{\ast}=1$, otherwise $\phi_{i}^{\ast}=0$. \( P \) is the probability matrix of classification. \(BCE\) stands for binary cross-entropy function. Then the server sends the model constructed above to the edge devices.

\subsubsection{Sparse Local Training}
The edge devices train the received model. In order to simplify and efficiently sparse the model, we choose channel-level sparsification, which makes a compromise between flexibility and realizability. The specific sparse training scheme is mainly divided into the following three steps.
\begin{itemize}
\item\textit{Selection of Scaling Factor:} First, inspired by \cite{liu2017learning}, we choose the $\gamma$ factor in the BN layer\cite{ioffe2015batch} as the scaling factor for sparse training. The BN layer refers to batch normalization, which is widely used in CNN networks to accelerate network training and convergence, preventing gradient loss and avoiding overfitting. BN is usually added after the convolution layer or linear layer. The specific process of BN conversion is as follows:

\begin{equation}
z=\frac{D_{\mathrm{in}}-\mu _{\mathcal{B}}}{\sqrt{\sigma _{\mathcal{B}}^{2}+\epsilon  } } ;\quad D_{\mathrm{out}}=\gamma z+\beta,
\end{equation}
where $D_{\mathrm{in}}$ and $D_{\mathrm{out}}$ are the input and output of BN layer, respectively. $\mu_{\mathcal{B}}$ and $\sigma_{\mathcal{B}}$ denote the average and standard deviation of input activations over the current small batch $\mathcal{B}$. $\gamma$ and $\beta$ are trainable transformation parameters, which can be used to linearly transform the normalized activation into any scale.
\item\textit{Sparse Training of Object Detection Model:} The scaling factor is used to sparse the object detection model, and a typical method is to multiply the scaling factor by the output of the corresponding channel. Then we can jointly train the network weight and scaling factors, with the L1 regularization\cite{tibshirani1996regression} imposed on the scaling factors. L1 regularization is a method to control the complexity of the model and reduce overfitting, which is generally used to achieve sparsity. Therefore, we add the L1 penalty term corresponding to the $\gamma$ factor after the loss function of the object detection model. The complete loss function is as follows:
\begin{equation}
\mathcal{L}=L+\lambda \sum_{\gamma \in \Gamma }g\left ( \gamma  \right ),
\end{equation}
where \(L\) represents the loss function of object detection in Formula (3). $g(\gamma)=|\gamma|$ corresponds to the sparse induction penalty of $\gamma$, and $\lambda$ is used to balance the two terms. 
\item\textit{Acquisition of Sparse Model:} After sparse training using L1 regularization, each edge device will obtain an object detection model with several scaling factors close to zero. Then, we can generate a binary mask \(m\) to sparse the model according to the global threshold, which is defined as a percentage of all scaling factors, equivalent to a $\gamma$ threshold. $m_{k}\in\left\{0,1\right\}^{d}$ represents the sparse binary mask of the \(k\)-th device. The element of \(m\) is 1 means that the weight of the corresponding position is retained, and 0 is discarded. Then, the model rounds off all the channels and input and output connections corresponding to the scaling factor lower than the threshold value. The specific operation is to set the value to zero at the corresponding position of the binary mask, and the left channel value is 1. The trained model multiplies with the binary mask to achieve sparsity. The formula is illustrated as follows:
\begin{equation}
\widetilde{w}_{k}^{t}=w_{k}^{t}\odot m_{k}^{t},
\end{equation}
where $m_{k}^{t}$ represents the sparse binary mask obtained after the \(k\)-th edge device sparse training in the \(t\)-th round of federated learning. $w_{k}^{t}$ is the sparse model weight obtained by multiplying the complete sparse model weight $\widetilde{w}_{k}^{t}$ with the binary mask.

\end{itemize}

\subsubsection{Federated Aggregation}
After the sparsification, different sparse models are obtained at each edge device. So the edge devices keep the model structure unchanged and only upload the remaining model updates to the server. Due to the diversity of model sparsity on different edge devices, the FedAvg algorithm cannot be directly applied in aggregation. Therefore, we improve the federated averaging algorithm in the next subsection.

\subsection{Federated Weight Aggregation Method}
In the federated learning framework, the FedAvg algorithm is widely used. However, in real situations, the computing resources of each edge device are different. In order to enable make these edge devices to work normally, we set different sparse rates on edge devices. Therefore, to guarantee the accuracy of global federated learning, we integrate the sparsity index into the aggregation algorithm. Specifically, the weight is determined according to the inverse ratio of the sparsity rate. For example, as shown in Figure 2, the sparsity ratio of the three devices is 40$\%$: 30$\%$: 20$\%$, and then the weight ratio of aggregation can be calculated as 3: 4: 6. The main idea is that the model with a high sparsity rate will have a relatively poor performance. Therefore, when aggregating, we allocate its weight to a lower aggregation proportion and set a higher weight proportion to the model with a low sparsity rate. 

After the aggregation is completed on the server side, the model is distributed to each edge device, and then multiplied by the binary mask \(m\) reserved by each edge device before the next round of training. The overall algorithm is summarized in Algorithm 1.

\begin{figure}[t]
    \centering
    \includegraphics[width=0.8\linewidth]{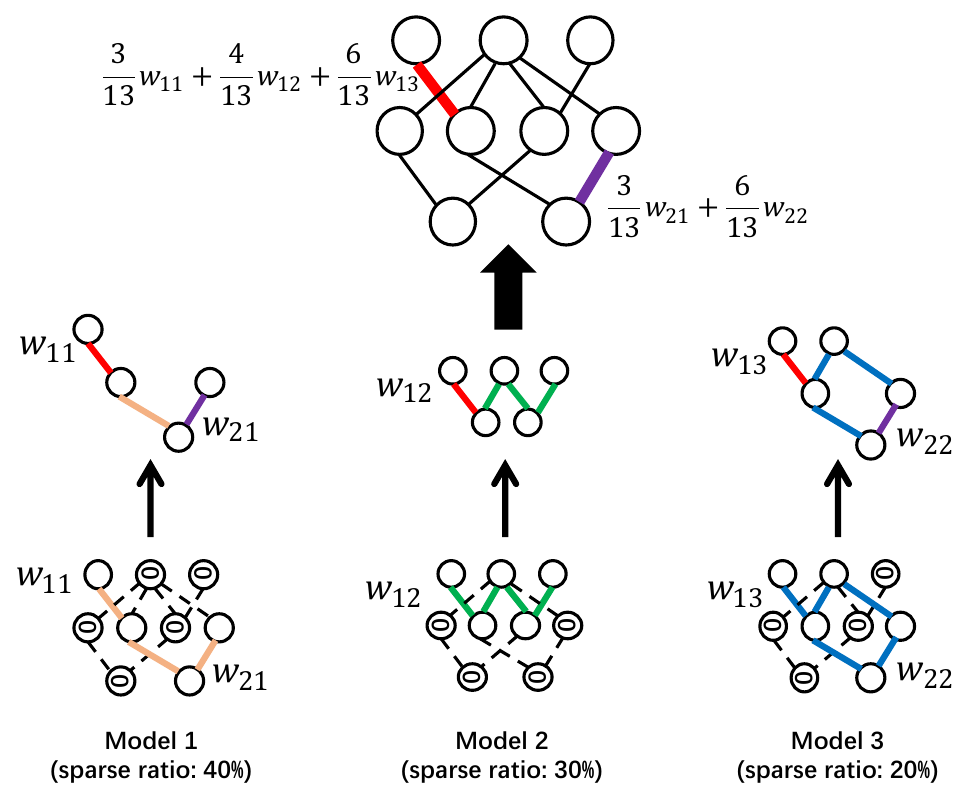}
    \caption{Schematic diagram of FedWeg aggregation method. (The sparsity ratio of the three devices is 40$\%$: 30$\%$: 20$\%$, then the weight ratio of aggregation is 3: 4: 6.)}
    \label{fig:my_label}
\end{figure}

\begin{algorithm}[htpb]
\caption{FedWeg Aggregation Algorithm}\label{alg:alg1}
\textbf{Server:} 
\begin{algorithmic}[1]
\STATE Initialize global model parameters $w^{0}$
\STATE {\textbf{for}} each round $t=1,2...$ {\textbf{do}}
\STATE \hspace{0.5cm}$\mathcal{K}_{t}\gets$(Sample a subset edge device from $\mathcal{K}$)
\STATE \hspace{0.5cm}{\textbf{for}} each edge device $k \in \mathcal{K}_{t}$ \textbf{in parallel do}
\STATE \hspace{1cm}$w_{k}^{t+1} \gets$ EdgeModelUpdate$(k,s_{k},w^t)$
\STATE \hspace{0.5cm}{\textbf{end for}}
\STATE \hspace{0.5cm}$w^{t+1}\gets \sum\limits_{k\in \mathcal{K}_{t}}\frac{\frac{1}{s_{k}}}{\sum_{k\in \mathcal{K}_{t}}\frac{1}{s_{k}}}\widetilde{w}_{k}^{t+1}$
\STATE \textbf{end for}\\ \textbf{EdgeModelUpdate$(k,s_{k},w)$:}
\STATE $\alpha $ is the learning rate; \(s\) is the sparse rate
\STATE $\mathcal{B}\gets$ (split into batches of size $B$)
\STATE {\textbf{for}} each epoch $e=1,...,E$ \textbf{do} 
\STATE \hspace{0.5cm} \textbf{for} batch $b\in \mathcal{B}$ \textbf{do}
\STATE \hspace{1cm}Sparse training:
\STATE \hspace{1cm}$w\gets w-\alpha\bigtriangledown \mathcal{L}\left(w,\gamma \right)$ 
\STATE {\textbf{end for}}
\STATE Set the threshold value according to \(s\)
\STATE Generate sparse binary mask \(m\) from Threshold
\STATE $\widetilde{w}\gets w\odot m$ 
\STATE return $\widetilde{w}$ to server
\end{algorithmic}
\label{alg1}
\end{algorithm}

Especially, the process of the server uses the FedWeg algorithm to obtain a new global model $w^{t+1}$ (Line 4-8) is described as follows:
\begin{equation}
w^{t+1}= \sum_{k\in \mathcal{K}_{t}}\frac{\frac{1}{s_{k}}}{\sum_{k\in \mathcal{K}_{t}}\frac{1}{s_{k}}}\widetilde{w}_{k}^{t},
\end{equation}
where $s_{k}$ is the sparsity rate of the \(k\)-th edge device, and all weights are aggregated in inverse proportion to the sparsity rate.

\section{Experiment}
In this section, we conducted in-depth experiments on real-life datasets to evaluate the accuracy and transmission overhead of the proposed scheme. 
\subsection{Dataset and Data Processing}
\begin{figure}[b]
    \centering
    \includegraphics[width=0.8\linewidth]{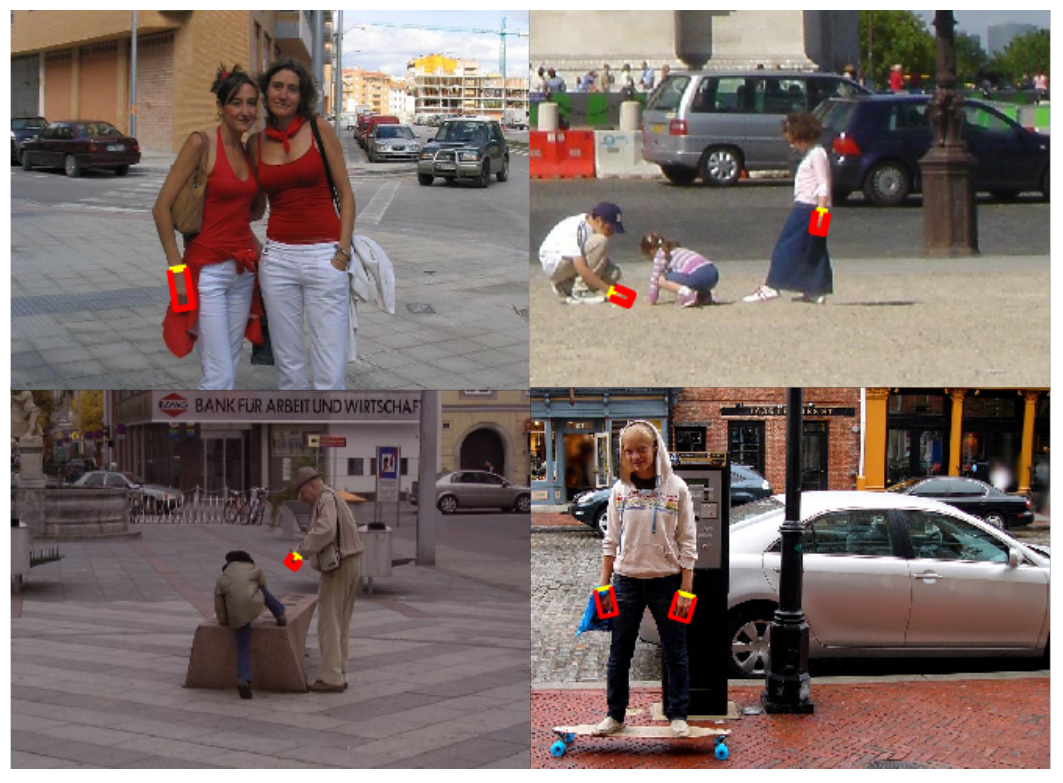}
    \caption{Example of images in the Hand Dataset}
\end{figure}

We select the Hand Dataset\cite{mittal2011hand} as the dataset used in the object detection experiments based on sparse federated training. Hand Dataset is a comprehensive dataset of hand images collected from different public image data sources. It contains 4807 training sets and 821 test sets. In collecting data, there are no restrictions on people's posture or visibility and the surrounding environment. The annotation in each image consists of a bounding rectangle, as shown in Figure 3.

\subsection{Experimental Design and Evaluation Criteria}

\paragraph{Object Detection using Pytorch Framework}

In the experiment, we use Python to train the YOLO model in the Pytorch framework. The framework can simply and quickly build a network by using the libraries. Our object detection model is derived based on YOLOv3, with a 416×416 input image size. After the model is built on the server side, the hyperparameters are adjusted to adapt to the model training.

\paragraph{Sparse Federated Learning Setup}

Our models are trained by a server equipped with NVIDIA RTX 2080Ti GPU and three laptops with limited computing resources. On the server side, we use the Ubuntu 18.04 operating system. After the YOLOv3 model initialization is completed on the server side, the model is distributed to the three laptops. The laptops use their private data set to train spare models. In terms of data distribution, each independent laptop would randomly obtain 2000 pictures as the training set. On this basis, we use three algorithms in our experiments, including the improved FedWeg algorithm based on sparse federated training (S-FedWeg), the FedAvg algorithm based on sparse federated learning (S-FedAvg) and the traditional non-sparse FedAvg algorithm (FedAvg). We set the sparsity rate of three laptops to 20$\%$, 30$\%$, and 40$\%$, respectively. In addition, for S-FedWeg and S-FedAvg, we set $\lambda =10^{-4}$, which can control the degree of sparsity. Each experiment simulates 15 communication rounds with 5 local epochs. We use the mAP index and the transmission overhead in the communication process to evaluate the performance. MAP (Mean Average Precision) is an index to measure the recognition accuracy in the object detection area. In these experiments, we express mAP as the Average Precision under a 0.5 IoU threshold.

\begin{figure}[h]
    \centering
    \includegraphics[width=\linewidth]{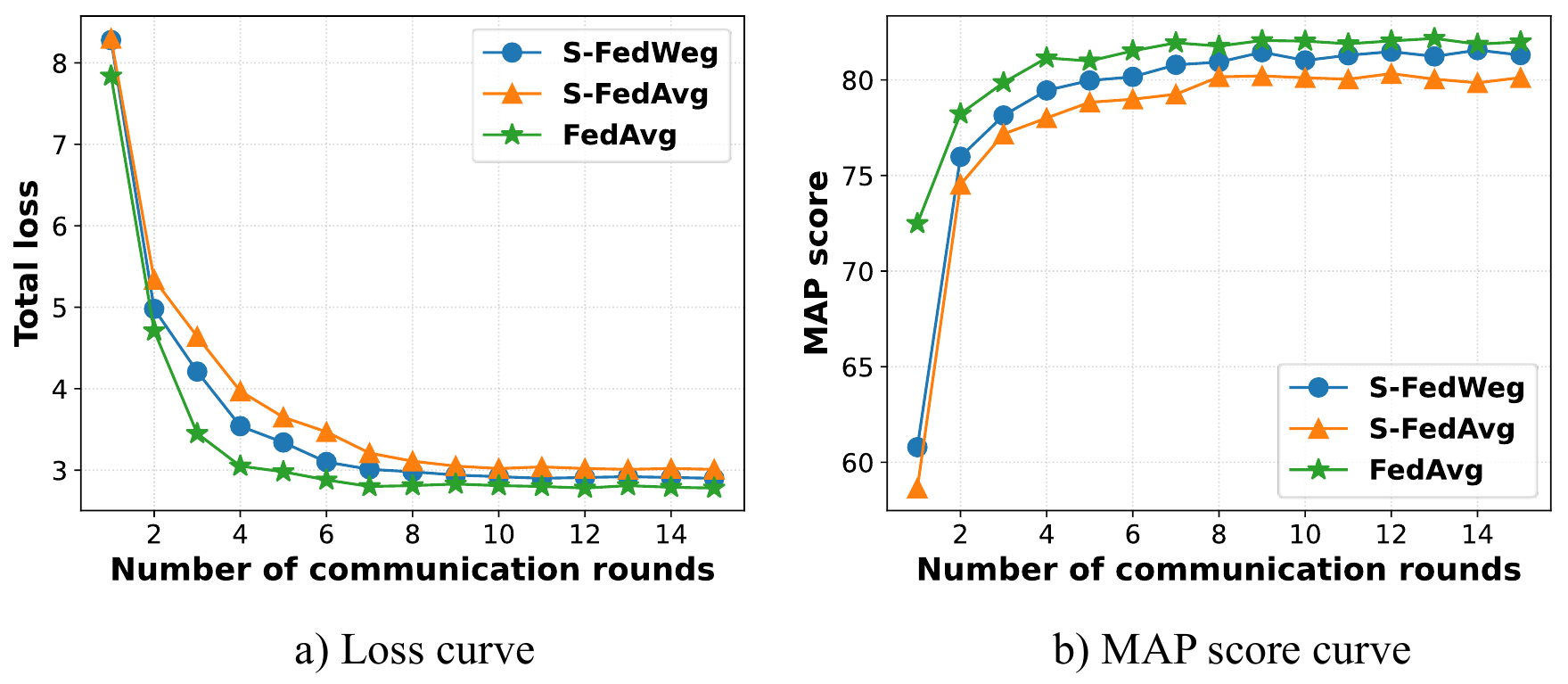}
    \caption{Comparison between S-FedWeg and the other two methods on model performance, where b) shows the mAP score of the proposed S-FedWeg is always higher than that of S-FedAvg.}
\end{figure}

\begin{table}[htbp]
\caption{Comparison between S-FedWeg and the other two methods on mAP and saved transmission bit}
\begin{center}
\begin{tabular}{|c|c|c|c|}
\hline
\textbf{Method} & \textbf{Data Size} & \textbf{MAP} & \multicolumn{1}{l|}{\textbf{Bit Saved(MB)}} \\ \hline
S-FedWeg        & 4807               & 81.30        & 3234.2                                      \\ \hline
S-FedAvg        & 4807               & 80.12        & 3177.3                                      \\ \hline
FedAvg          & 4807               & 81.99        & 0                                           \\ \hline
\end{tabular}
\end{center}
\end{table}

\subsection{Results and Analysis}
We compare S-FedWeg with S-FedAvg and FedAvg in terms of performance-communication tradeoff. The results in Figure 4 and Table \uppercase\expandafter{\romannumeral1} show that S-FedWeg can achieve high mean average accuracy with a significant reduction in communication overhead. 

First, the purpose of L1 regularization is to make several scaling factors close to 0, and $\lambda$ in Equation 8 is a parameter that affects the degree of sparsity. Therefore, in Figure 5, we draw the distribution of scaling factors of networks with different $\lambda$ values. And we find that when $\lambda=10^{-5}$, the scaling factor is not large enough, and when $\lambda=10^{-3}$, the scaling factor is mostly close to 0, which may leads to poor performance of the aggregation model in the subsequent federated learning process. Therefore, $\lambda=10^{-4}$ is selected.

Second, compared to FedAvg, S-FedWeg has a negligible performance loss. Especially, the mAP only has a 0.69 drop in the 15th round, which shows the practicality of the proposed sparse training design. In addition, a total of 3234.2MB are saved during the 15 rounds of transmission, which shows the great advantages of alleviating the transmission burdens.

Third, compared to S-FedAvg, S-FedWeg improves the model performance to a certain extent. Especially, Figure 4b) shows that during 15 rounds of communication, the mAP value of S-FedWeg is always higher than that of S-FedAvg. In addition, the transmission bits saved by FedWeg increased slightly, by 56.9MB.

Fourth, we further conduct experiments for S-FedWeg on different numbers of local epochs. It can be seen from Figure 6 that when the number of epochs of each communication round increases, the performance of the model improves. However, after the aggregation model reaches the mAP score at around 82, the performance comes to converge.

\begin{figure*}[t]
    \centering
    \includegraphics[width=0.8\textwidth]{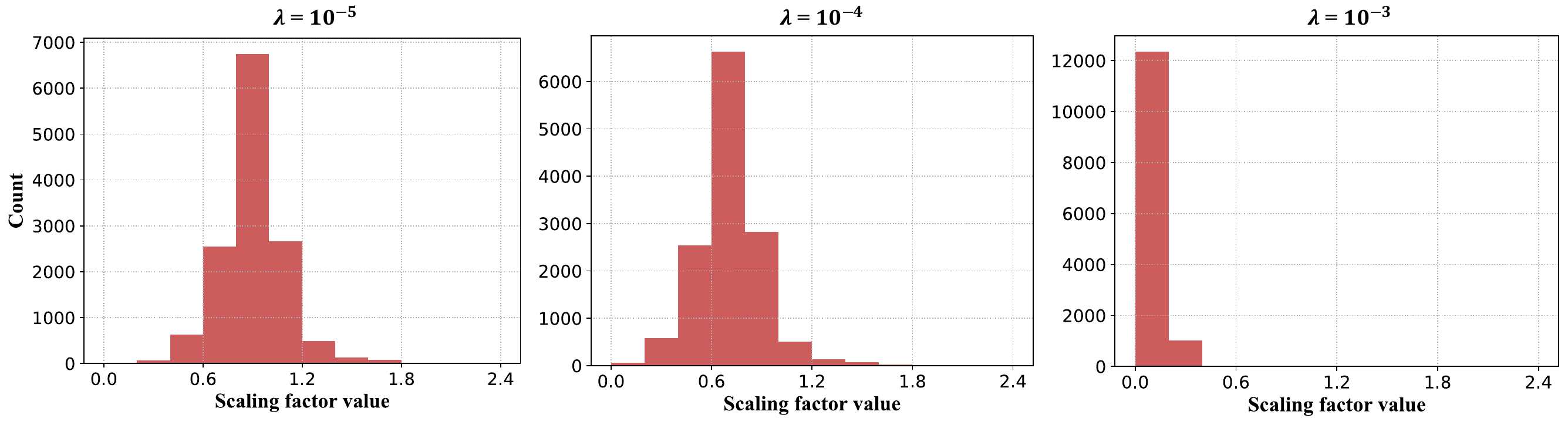}
    \caption{Distributions of scaling factors under various degrees of sparsity. With the increase of $\lambda$, scaling factors become sparser.}
\end{figure*}

\begin{figure*}[t]
    \centering
    \includegraphics[width=0.8\textwidth]{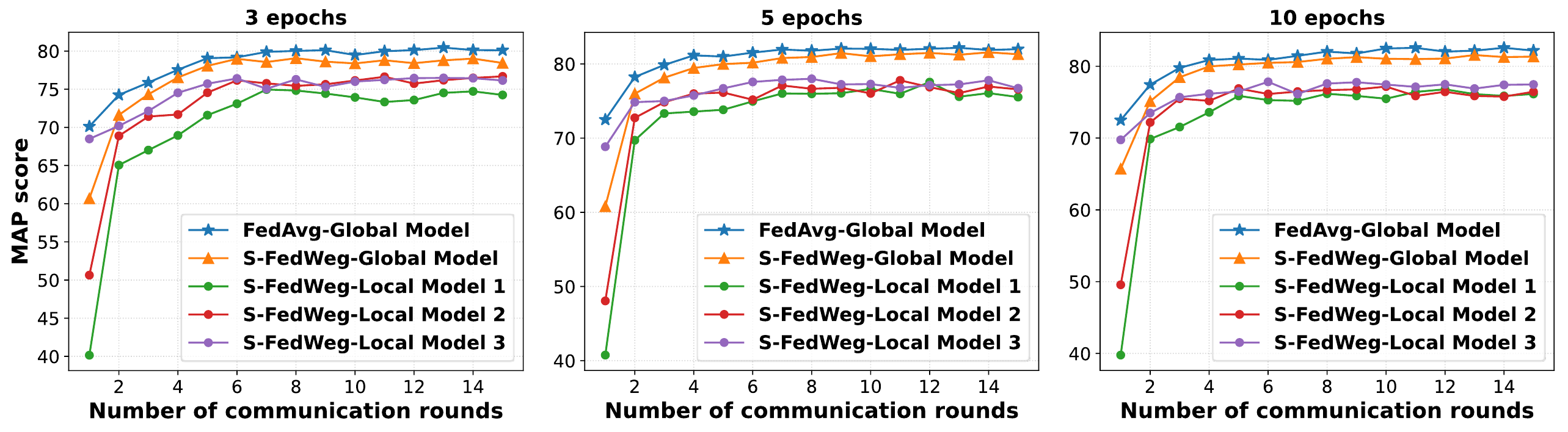}
    \caption{The impact of different epochs on the model performance. With the increase in the number of epochs, the model performance improves.}
\end{figure*}

\section{Conclusion}
In this paper, we have proposed an object detection scheme based on sparse federated training. This scheme uses local data to train local models on edge devices, and only lightweight local parameters are uploaded to the server for aggregation, thus avoiding direct data sharing and reducing communication costs. In addition, considering the various sparsity rates of models on different edge devices, a weight aggregation method based on the inverse ratio of sparsity rates has been proposed. To verify the effectiveness of the proposed algorithm, we have conducted real-life experiments under the framework of YOLOv3, and 30.20$\%$ overhead can be saved under a similar detection performance of the non-sparsity-based algorithm.

\bibliographystyle{ieeetr}

\begin{thebibliography}{10}

\bibitem{chen2020edge}
C.~Chen, B.~Liu, S.~Wan, P.~Qiao, and Q.~Pei, ``An edge traffic flow detection
  scheme based on deep learning in an intelligent transportation system,'' {\em
  IEEE Transactions on Intelligent Transportation Systems}, vol.~22, no.~3,
  pp.~1840--1852, 2021.

\bibitem{cheng2020accessibility}
J.~Cheng, G.~Yuan, M.~Zhou, S.~Gao, C.~Liu, H.~Duan, and Q.~Zeng,
  ``Accessibility analysis and modeling for iov in an urban scene,'' {\em IEEE
  Transactions on Vehicular Technology}, vol.~69, no.~4, pp.~4246--4256, 2020.

\bibitem{konevcny2016federated}
J.~Kone{\v{c}}n{\`y}, H.~B. McMahan, D.~Ramage, and P.~Richt{\'a}rik,
  ``Federated optimization: Distributed machine learning for on-device
  intelligence,'' {\em arXiv preprint arXiv:1610.02527}, 2016.

\bibitem{li2020federated}
T.~Li, A.~K. Sahu, A.~Talwalkar, and V.~Smith, ``Federated learning:
  Challenges, methods, and future directions,'' {\em IEEE Signal Processing
  Magazine}, vol.~37, no.~3, pp.~50--60, 2020.

\bibitem{han2015deep}
S.~Han, H.~Mao, and W.~J. Dally, ``Deep compression: Compressing deep neural
  networks with pruning, trained quantization and huffman coding,'' {\em arXiv
  preprint arXiv:1510.00149}, 2015.

\bibitem{hinton2015distilling}
G.~Hinton, O.~Vinyals, J.~Dean, {\em et~al.}, ``Distilling the knowledge in a
  neural network,'' {\em arXiv preprint arXiv:1503.02531}, vol.~2, no.~7, 2015.

\bibitem{liu2017learning}
Z.~Liu, J.~Li, Z.~Shen, G.~Huang, S.~Yan, and C.~Zhang, ``Learning efficient
  convolutional networks through network slimming,'' in {\em 2017 IEEE
  International Conference on Computer Vision (ICCV)}, pp.~2755--2763, 2017.

\bibitem{peng2012rasl}
Y.~Peng, A.~Ganesh, J.~Wright, W.~Xu, and Y.~Ma, ``Rasl: Robust alignment by
  sparse and low-rank decomposition for linearly correlated images,'' {\em IEEE
  Transactions on Pattern Analysis and Machine Intelligence}, vol.~34, no.~11,
  pp.~2233--2246, 2012.

\bibitem{DBLP:journals/corr/abs-2108-13323}
C.~Wu, F.~Wu, R.~Liu, L.~Lyu, Y.~Huang, and X.~Xie, ``Fedkd: Communication
  efficient federated learning via knowledge distillation,'' {\em CoRR},
  vol.~abs/2108.13323, 2021.

\bibitem{DBLP:conf/sensys/0005SZZLC21}
A.~Li, J.~Sun, X.~Zeng, M.~Zhang, H.~Li, and Y.~Chen, ``Fedmask: Joint
  computation and communication-efficient personalized federated learning via
  heterogeneous masking,'' in {\em Proceedings of the 19th ACM Conference on
  Embedded Networked Sensor Systems}, pp.~42--55, 2021.

\bibitem{dai2022dispfl}
R.~Dai, L.~Shen, F.~He, X.~Tian, and D.~Tao, ``Dispfl: Towards
  communication-efficient personalized federated learning via decentralized
  sparse training,'' {\em arXiv preprint arXiv:2206.00187}, 2022.

\bibitem{mcmahan2017communication}
B.~McMahan, E.~Moore, D.~Ramage, S.~Hampson, and B.~A. y~Arcas,
  ``Communication-efficient learning of deep networks from decentralized
  data,'' in {\em Artificial Intelligence and Statistics}, pp.~1273--1282,
  PMLR, 2017.

\bibitem{1}
J.~Redmon, S.~Divvala, R.~Girshick, and A.~Farhadi, ``You only look once:
  Unified, real-time object detection,'' in {\em 2016 IEEE Conference on
  Computer Vision and Pattern Recognition (CVPR)}, pp.~779--788, 2016.

\bibitem{redmon2018yolov3}
J.~Redmon and A.~Farhadi, ``Yolov3: An incremental improvement,'' {\em arXiv
  preprint arXiv:1804.02767}, 2018.

\bibitem{ioffe2015batch}
S.~Ioffe and C.~Szegedy, ``Batch normalization: Accelerating deep network
  training by reducing internal covariate shift,'' in {\em International
  Conference on Machine Learning}, pp.~448--456, PMLR, 2015.

\bibitem{tibshirani1996regression}
R.~Tibshirani, ``Regression shrinkage and selection via the lasso,'' {\em
  Journal of the Royal Statistical Society: Series B (Methodological)},
  vol.~58, no.~1, pp.~267--288, 1996.

\bibitem{mittal2011hand}
A.~Mittal, A.~Zisserman, and P.~H. Torr, ``Hand detection using multiple
  proposals.,'' in {\em Bmvc}, vol.~2, p.~5, 2011.

\end{thebibliography}

\end{document}